\documentclass[11pt]{article}

\usepackage{acl}

\usepackage{times}
\usepackage{latexsym}
\usepackage{amssymb}
\usepackage{acl}
\usepackage[T1]{fontenc}

\usepackage[utf8]{inputenc}

\usepackage{microtype}

\usepackage{inconsolata}

\usepackage{algorithm}
\usepackage{algorithmic}
\usepackage{amsmath}
\usepackage{graphicx}
\usepackage[table,xcdraw]{xcolor}
\usepackage{diagbox}
\usepackage{multirow}
\usepackage{tabularx}
\usepackage{xcolor}
\newcolumntype{Y}{>{\centering\arraybackslash}X}
\usepackage{amsfonts,amssymb}
\usepackage{booktabs}
\usepackage{makecell}

\usepackage{graphicx}

\usepackage{amsmath}

\title{GaLa: Hypergraph-Guided Visual Language Models for Procedural Planning}

\author{
Kun Wang$^{1,3}$\footnotemark[2],
Yiming Li$^{1,3}$\footnotemark[2],
Mingcheng Qu$^{1,3}$\footnotemark[2],
Aqiang Zhang$^{1,3}$,
Guang Yang$^{1}$,
Tonghua Su$^{1,2,3}$\footnotemark[1]
\\
$^{1}$ Harbin Institute of Technology, Harbin, China \\
$^{2}$ Guangdong Laboratory of Artificial Intelligence and Digital Economy (SZ), Shenzhen, China \\
$^{3}$ Chongqing Research Institute of HIT, Chongqing, China
}

\begin{document}
\maketitle

\begingroup
\renewcommand{\thefootnote}{\fnsymbol{footnote}}
\footnotetext[2]{Equal contribution.}
\footnotetext[1]{Corresponding author.}
\endgroup

\begin{abstract}
Implicit spatial relations and deep semantic structures encoded in object attributes are crucial for procedural planning in embodied AI systems. However, existing approaches often over-rely on the reasoning capabilities of vision language models (VLMs) themselves, while overlooking the rich structured semantic information that can be mined from multimodal inputs. As a result, models struggle to effectively understand functional spatial relationships in complex scenes. To fully exploit implicit spatial relations and deep semantic structures in multimodal data, we propose GaLa, a vision–language framework for multimodal procedural planning. GaLa introduces a hypergraph-based representation, where object instances in the image are modeled as nodes, and region-level hyperedges are constructed by aggregating objects according to their attributes and functional semantics. This design explicitly captures implicit semantic relations among objects as well as the hierarchical organization of functional regions. Furthermore, we design a Tri-View HyperGraph Encoder that enforces semantic consistency across the node view, area view, and node–area association view via contrastive learning, enabling hypergraph semantics to be more effectively injected into downstream VLM reasoning. Extensive experiments on the ActPlan-1K and ALFRED benchmarks demonstrate that GaLa significantly outperforms existing methods in terms of execution success rate, LCS, and planning correctness.

\end{abstract}

\section{Introduction}

In recent years, research in procedural planning has significantly advanced the capability of embodied agents to execute complex instructions by understanding multimodal information~\cite{wan2025reca, li2023behavior,sun2025llapa}. Procedural planning refers to the process of progressively decomposing high-level linguistic commands into a sequence of precise actions required to accomplish a task~\cite{wang2025procworld}, ultimately enabling embodied agents to perform task planning and execution in real-world environments~\cite{huang2025crmarena}. During this process, textual instructions enable the agent to effectively understand the required task~\cite{zhao2024see}, while visual inputs allow the agent to perceive its surroundings with precision~\cite{lin2025showui}. Through multimodal fusion~\cite{yang2024embodied}, the embodied agent is ultimately able to execute the specified instructions.

\begin{figure*}[t]
  \includegraphics[width=\linewidth]{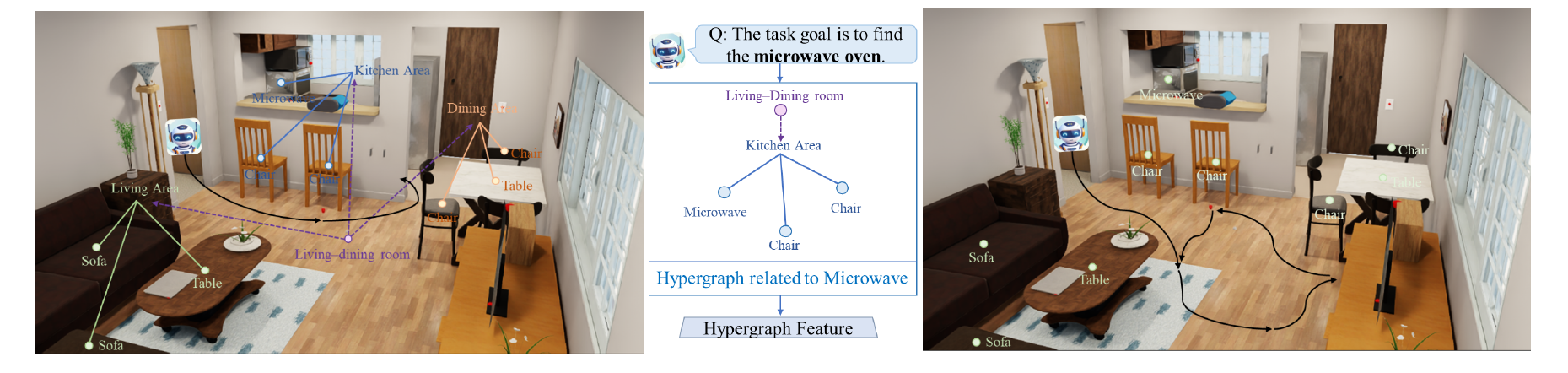}
  \caption{On the left, when the hypergraph is introduced, the deep semantic information contained in the visual data of the room is extracted, enabling the execution of the procedural planning process. In the middle, the abstraction process of building the hypergraph is shown. On the right, without the introduction of the hypergraph, the procedural planning results in a logical deadlock.}
  \label{fig:example}
\end{figure*}

Recent research in procedural planning can be categorized into two main streams based on the modality they primarily rely on: Large Language Models (LLMs)~\cite{guan2023leveraging, silver2024generalized, li2024embodied} and Vision-Language Models (VLMs)~\cite{mu2023embodiedgpt, zhai2024fine, yang2025embodiedbench}. VLM-based methods, by contrast, integrate visual observations with language input, enabling improved spatial perception and grounding. This capability allows VLMs~\cite{yang2026chain} to generate action plans that better respect physical constraints in the environment. Despite this advantage, existing VLM-based approaches predominantly operate on explicit spatial representations, such as bounding boxes or object coordinates, and focus primarily on image-level semantic cues. As a result, they remain limited in capturing deeper semantic structures that are implicitly embedded in visual scenes.
In particular, two critical limitations emerge. First, implicit spatial relations conveyed through object attributes—such as "can-be-placed-on", "is-supported-by" are often overlooked, as they are not explicitly encoded in geometric representations. Second, hierarchical semantics arising from groups of objects forming functional regions (e.g., "dining area", "kitchen area") are rarely modeled in a structured manner. Consequently, the agent’s understanding of the scene is reduced to a collection of isolated objects, which can lead to suboptimal or even degenerate planning behaviors, such as repetitive or oscillatory actions in ambiguous contexts (as illustrated in Figure~\ref{fig:example}).

This phenomenon raises a critical question: Are current approaches overly reliant on the explicit spatial reasoning capabilities of VLMs, while neglecting the opportunity to mine deeper semantic structures from images?

To address this challenge, we propose GaLa, a framework that introduces a hypergraph-based representation as an intermediate semantic abstraction between vision and language. Rather than treating spatial reasoning as purely geometric, GaLa models the scene as a hypergraph that explicitly captures both object-level semantic attributes and area-level functional organization. Specifically, GaLa constructs a semantic hypergraph in which nodes correspond to individual object instances, while hyperedges represent functional or spatial regions inferred from object attributes (e.g., "dining area", "kitchen area"). By aggregating object-level semantics into structured area-level representations, the hypergraph elevates
implicit spatial relations—originally latent in textual and semantic cues—into an explicit, reasoning-friendly form. To effectively integrate this structured knowledge into downstream reasoning, we further introduce a "Tri-View HyperGraph Encoder", which employs contrastive learning to enforce semantic consistency across three complementary views: object-level (node-view), region-level (area-view), and object–region association (all-view). This tri-view objective encourages the model to learn robust representations that preserve both fine-grained object semantics and higher-order relational structure, enabling more coherent and spatially grounded procedural planning. Through this design, GaLa provides VLMs with richer semantic context and mitigates common planning failures arising from ambiguous or under-structured scene representations.

Experimental validation on the ActPlan-1k benchmark dataset shows that the GaLa model achieves significant performance improvements. The results indicate that, thanks to the explicit modeling of implicit spatial relations and object cluster semantics via the hypergraph, our framework can guide the VLM to generate action sequences that are more spatially accurate and semantically consistent, demonstrating superior performance in complex procedural planning tasks.


Our key contributions are summarized as follows:

• We propose GaLa, the first framework that introduces graph-theoretic structural semantic information into a VLM-based procedural planning architecture.


• We design a HyperGraph Semantic Encoder that models visual information to construct a hypergraph enriched with semantic information, thereby enhancing the structural semantics inherent in the image.


• We introduce a Tri-View HyperGraph Encoder that employs a contrastive learning approach to ensure that the hypergraph information is fully preserved and effectively transmitted into the VLM.



\begin{figure*}[t]
  \includegraphics[width=\linewidth]{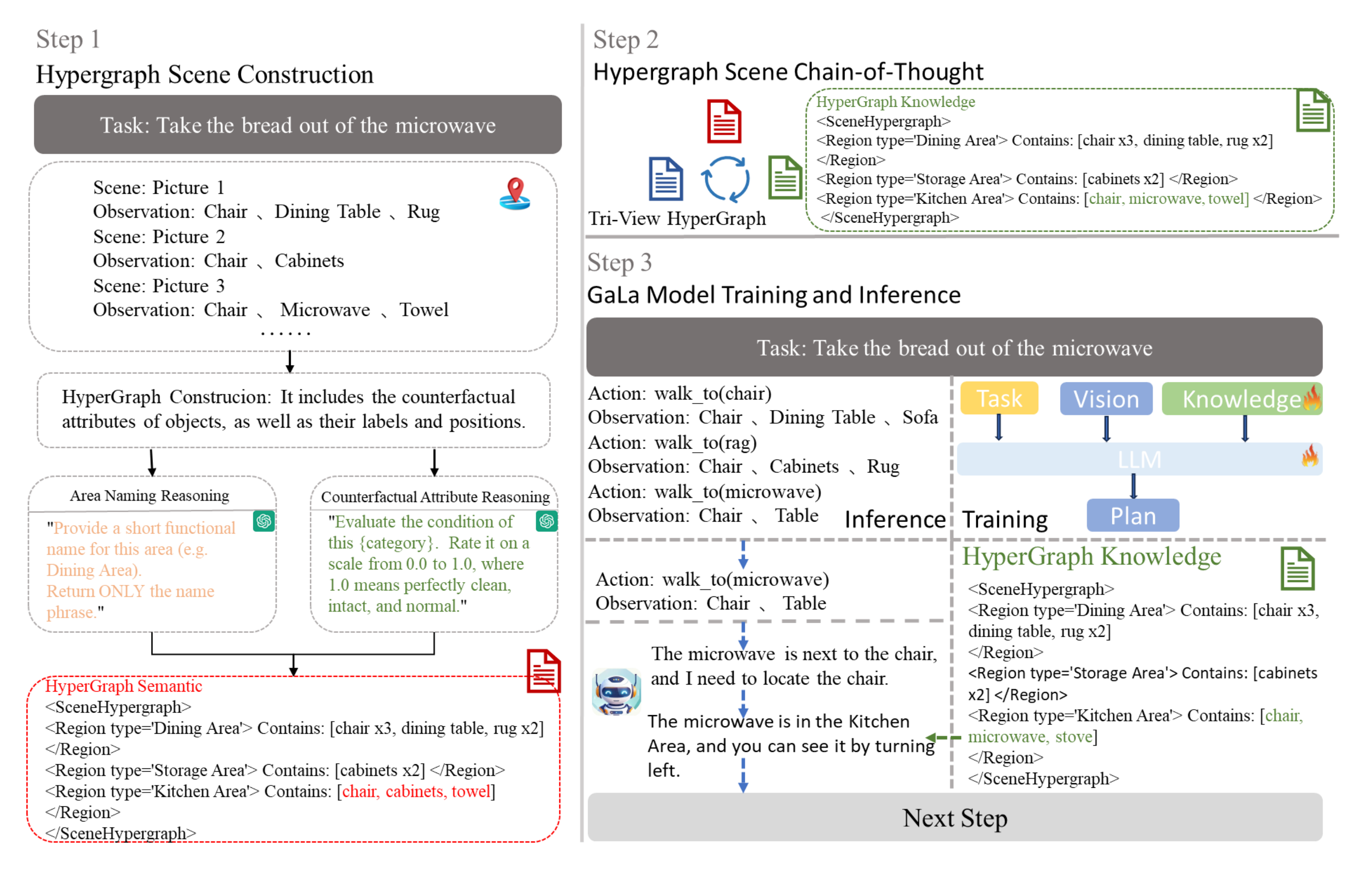}
  \caption{We present the model pipeline for GaLa. In $\textit{Step 1}$, we initially build the hypergraph semantic information. In $\textit{Step 2}$, we optimize the hypergraph information for CoT using the semantic information from $\textit{Step 1}$ and the Tri-View HyperGraph. In $\textit{Step 3}$, we train the constructed knowledge, enabling the model to predict the next action more accurately.}
  \label{fig:pipeline}
\end{figure*}

\begin{figure*}[t]
  \includegraphics[width=\linewidth]{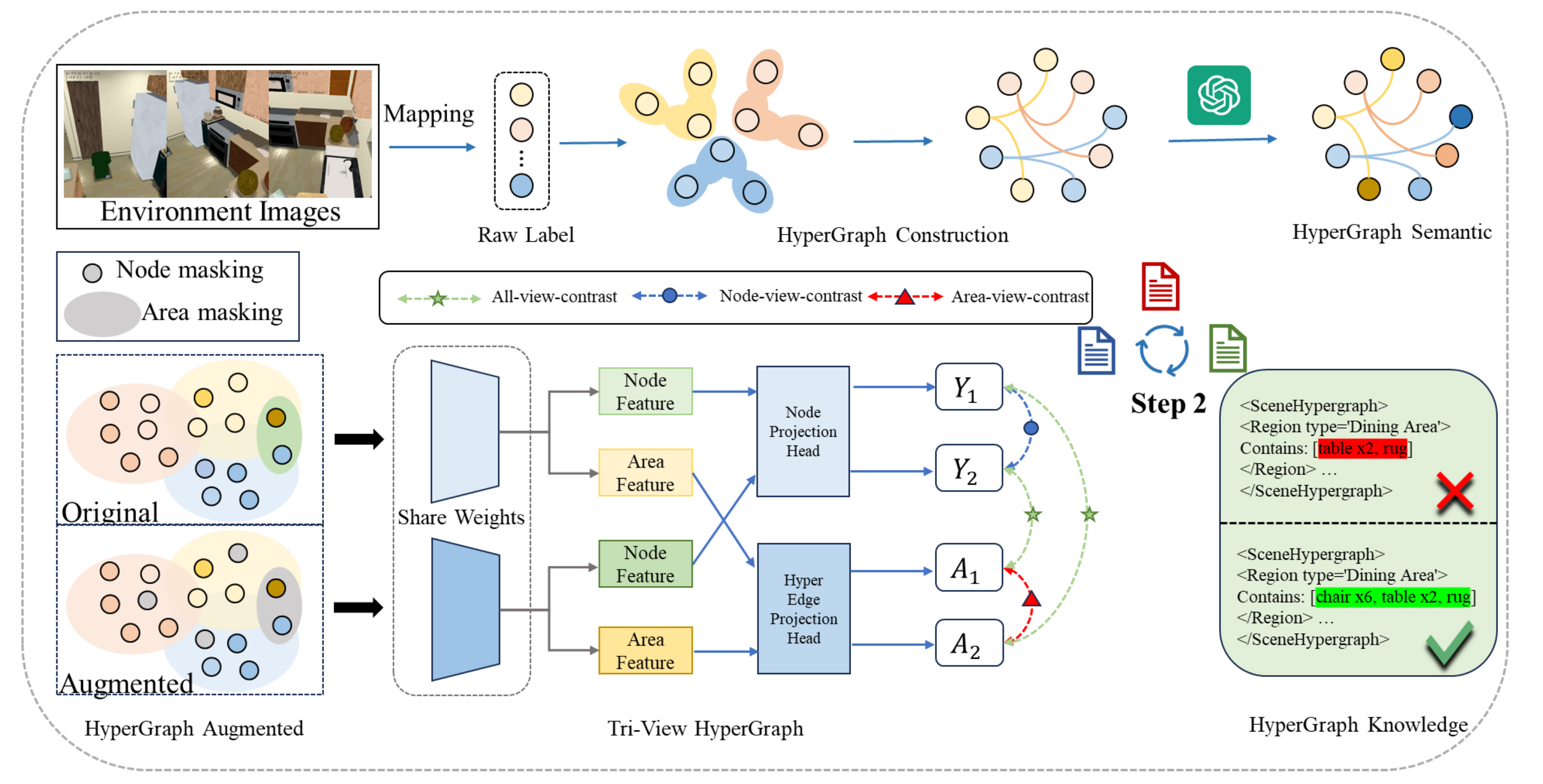}
  \caption{The detailed architecture of Step 2 is illustrated in Figure 2. The upper part depicts the process of Hypergraph Semantic Construction, while the lower-left part shows the Tri-View HyperGraph module, and the lower-right part presents the resulting HyperGraph Knowledges. The entire pipeline is organized as a chain-of-thought (CoT) process, which progressively refines and optimizes the HyperGraph Knowledges.}
  \label{fig:structuon}
\end{figure*}

\section{Related Work}

\subsection{Vision-Language Models (VLMs)}

VLMs jointly model visual and linguistic information and are widely used in multimodal tasks such as image captioning~\cite{cheng-etal-2025-caparena}, visual question answering~\cite{bhat2025expertneurons,lim2025vlr}, and image--text retrieval~\cite{liu2024mmbench,chen2015microsoft,plummer2015flickr30k}. Early contrastive methods, including CLIP~\cite{radford2021learning} and ALIGN~\cite{jia2021scaling}, established large-scale image--text alignment. 

Modern VLMs integrate visual features into LLM pipelines with powerful backbones (GPT-style, LLaMA-family and Qwen-series~\cite{yang2025qwen3}) and visual instruction tuning. Representative methods include Flamingo~\cite{alayrac2022flamingo}, BLIP-2~\cite{li2023blip}, and LLaVA~\cite{liu2023visual}, using different mechanisms to inject visual information. Despite strong performance, most VLMs treat images as unstructured tokens, limiting explicit modeling of object relations and higher-order structure, motivating structured visual representations.



\subsection{Procedural Planning.}




Procedural planning decomposes high-level goals into executable action sequences. Early symbolic planners with manually defined state and action spaces often struggle to generalize to perceptually rich, open-world environments~\cite{lu2022neuro,srivastava2014combined}.

Recent large language models support learning-based planning from natural language, using chain-of-thought reasoning, tool invocation, and iterative refinement~\cite{wei2022chain,schick2023toolformer,shinn2023reflexion}. VLM-based planners further ground language reasoning in visual observations. Benchmarks like ActPlan-1K~\cite{su2024actplan} show that integrating vision improves plan feasibility and coherence.

Most methods encode visual inputs implicitly, lacking explicit modeling of object-level structure and spatial relations, motivating structured semantic abstractions for more interpretable and reliable VLM-based planning.

\subsection{Hypergraph-based Contrastive Learning}

Hypergraph-based contrastive learning captures higher-order relationships beyond pairwise connections. Recent work introduces a critical node‑aware hypergraph contrastive learning method~\cite{li2025critical} and applies similar ideas to multi-interest fairness in recommender systems~\cite{zheng-etal-2025-multi}. These studies demonstrate that explicitly modeling high-order structure can enrich representations, which inspires our HyperGraph Semantic Encoder and Tri-View HyperGraph Encoder to capture semantic structures in images and preserve them within VLM-based procedural planning.


\section{Method}
\subsection{Overview}
\paragraph{Problem Definition.} The workflow for performing procedural planning using a multimodal model is described as follows. We define the visual features as $I = \{i_{1}, i_{2},...,i_{n}\}$, which represent the environment perceived during the exploration process. The textual features are defined as $L = \{l_{goal}, l_{1},..., l_{m}\}$, where $l_{goal}$ denotes the execution goal of the task, and the remaining elements represent semantic guidance instructions required for task execution. The generated execution objective is defined as $A = \{a_{1}, a_{2},..., a_{t}\}$, which represents the sequence of actions required to accomplish the task. 


\paragraph{GaLa Pipeline.} The overall pipeline of GaLa can be summarized into three main steps, as illustrated in Figure~\ref{fig:pipeline}. \textbf{Step.1 : Hypergraph Semantic Construction.} Object nodes $x\in \mathbb{R}$ are first generated from visual images $v\in \mathbb{R}^{W \times H \times 3}$ obtained in a simulated environment. Each object node $x$ contains its position $p\in \mathbb{R}^{W \times H}$, semantic label $c \in \mathbb{R}$, and a textual description of its attributes $s \in \mathbb{R}$. The object labels $c$ are treated as graph nodes, while corresponding Area attributes $z \in \mathbb{R}$ generated by GPT from these labels are used as hyperedges $C$ to construct the hypergraph $\mathcal{G}$. In addition, a large language model is employed to score object attributes  $s$ to determine whether they correspond to normal or counterfactual attributes $r\in [0, 1]$. This process ultimately produces the hypergraph semantic information $\mathcal{G}$. \textbf{Step.2 : Hypergraph Scene Chain-of Thought.} The overall architecture of this step is depicted in Figure~\ref{fig:structuon}. The hypergraph semantic information  $\mathcal{G}$ generated in Step 1, together with the Tri-View Hypergraph Encoder $\mathcal{G}_{t}$ and Hypergraph Knowledge $\mathcal{G}_{k}$, forms a chain-of-thought reasoning process to progressively refine the hypergraph semantics. This step results in a comprehensive hypergraph knowledge $\mathcal{G}_{k}$ representation of the visual scene. \textbf{Step.3 : GaLa Training and Inference with Hypergraph Knowledge.} The hypergraph knowledge $\mathcal{G}_{k}$ produced in Step 2, along with visual inputs $v$ and task information $l$, is fed into the VLM for training. During inference, the next action $a$ is predicted based on the visual input, task instruction, and the hypergraph knowledge distilled from the scene. 
This structured knowledge enhances the textual representation by incorporating explicit spatial and semantic relationships, enabling the model to better capture the underlying scene structure. 
As a result, the model can generate actions with improved spatial awareness and avoid logical inconsistencies, leading to more coherent and reasonable action sequences.


\subsection{HyperGraph Semantic Encoder}

The HyperGraph Semantic Encoder module is illustrated in the upper part of Figure 3. We first obtain a set of images from the simulated environment $V = \{v_{1}, v_{2},...,v_{|M|}\},$
where each image $v_{i} \in \mathbb{R}^{W \times H \times 3}$. We apply the YOLO-World model to infer a set of objects, denoted as: $\mathcal{X} = \{x_{1}, x_{2},..., x_{|N|}\}$. Each object instance $x_{i}$ is represented by a tuple $x_{i} = (p_{i}, c_{i}, s_{i})$, where $p_{i} \in \mathbb{R}^{W \times H}$ denotes the spatial position of the object, $c_{i}$ it its category label and $s_{i}$ is a textual description of its attributes. To capture area-level spatial structure, we perform density-based spatial clustering over object positions. Specifically, we apply a DBSCAN-based~\cite{ester1996density} clustering algorithm to the set of object coordinates $\{p_{i}\}_{i=1}^{|N|}$, yielding a collection of spatial clusters:

\begin{equation}
       \mathcal{C} = D(\{p_{i}\}_{i=1}^{|N|},\epsilon, MinPts),
\end{equation}
where $\mathcal{C} = \{C_{1}, C_{2},...,C_{K}\}$, each cluster $C_{k}$ corresponds to a group of spatially coherent objects and is defined as:
\begin{equation}
       C_{k} = \{ x_{i} \in  \mathcal{X} | f(x_{i}) = k \},
\end{equation}
where $f:\mathcal{X} \rightarrow \{1,\ldots,K\}$ denotes the cluster assignment function. Finally, we construct a hypergraph by treating each object instance $x_{i}$ as a node and each spatial cluster $C_{k}$ as a hyperedge, resulting in the hypergraph $\mathcal{G}_{F} = (\mathcal{X},\mathcal{C})$. 
To enhance the extraction of semantic and spatial structures from visual scenes, we perform area-level semantic reasoning over the constructed hypergraph. For each hyperedge $C_{i}$, we collect the category labels of its constituent object nodes:
\begin{equation}
       L_{i} = \{c_{j}| x_{j} = (p_{j}, c_{j}, s_{j}) \in C_{i}\},
\end{equation}
where $c_{j}$ denotes the catagory label of object $x_{j}$. 
We feed the aggregated category labels $\mathcal{L}_{i}$ into the large language model $\mathcal{M}_{LLM}$ (InternVL-3) with a task-specific prompt to infer an area-level semantic label: $z_{i} = \mathcal{M}_{LLM}(\mathcal{L}_{i}; \mathcal{P}),$
where $\mathcal{P}$ denotes the prompt used to instruct the LLM for area semantic reasoning, and $z_{i}$ represents the area-level semantic label associated with hyperedge $C_{i}$ (e.g., Kitchen Area, Dining Area). \textbf{Counterfactual Attribute Scoring with LLM.} For each detected object instance $x_{j}$, we further assess whether its attribute description corresponds to a normal or counterfactual state.
We leverage the large language model $\mathcal{M}_{LLM}$ (InternVL-3) with a task-specific prompt $\mathcal{P}_{cf}$ to assign a counterfactual score to each attribute: $r_{j} = \mathcal{M}_{LLM}(s_{j}; \mathcal{P}_{cf}), r_{j} \in [0, 1].$



Here, $r_{j}$ approaching 0 indicates that the attribute is more likely to be normal, while values closer to 1 indicate a higher degree of abnormality. Through the above process, we obtain the hypergraph semantics of the visual information.

\subsection{Tri-View HyperGraph}
Through the above procedure, we obtain a hypergraph $\mathcal{G}_{F} = (\mathcal{X}, \mathcal{C})$. The corresponding incidence matrix is defined as $\mathbf{H} \in \{0,1\}^{|\mathcal{X}|\times|\mathcal{C}|}$. We then apply random masking to both hypergraph nodes and hyperedges. Specifically, the masked node- and hyperedge-level textual representations are given by:
\begin{equation}
     c_{i}^{(k)}=Mask^{(k)}(c_{i}), z_{j}^{(k)}=Mask^{(k)}(z_{i}).
\end{equation}
Based on this formulation, both the constructed hypergraph and its augmented variants can be derived as follows:

\begin{equation}
\begin{aligned}
     \mathcal{G}_{F}^{(1)} = (\mathcal{X}, \mathcal{C}, \mathbf{H}^{(1)}, c_{i}^{(1)}, z_{j}^{(1)}), \\
     \mathcal{G}_{F}^{(2)} = (\mathcal{X}, \mathcal{C}, \mathbf{H}^{(2)}, c_{i}^{(2)}, z_{j}^{(2)}),
\end{aligned}
\end{equation}
where the augmented incidence matrices are defined as:
\begin{equation}
     \mathbf{H}^{(k)} = \mathbf{M}^{(k)} \odot \mathbf{H}, k \in \{1,2\},
\end{equation}
$\mathbf{M}^{(k)}$ denotes the all-view masking matrix. After applying data augmentation, we employ the text encoder of the InternVL model to separately encode node-level texts and hyperedge-level texts, obtaining the corresponding node and hyperedge representations.

\begin{equation}
\begin{aligned}
    P^{(k)} = \mathcal{E}_{\theta} (\{ c_{i}^{(k)} \}_{i=1}^{N})\in \mathbb{R}^{N \times d}, \\
    Q^{(k)} = \mathcal{E}_{\theta} (\{ z_{j}^{(k)} \}_{j=1}^{K})\in \mathbb{R}^{K \times d}.    
\end{aligned}
\end{equation}

To decouple semantic encoding from contrastive optimization, we employ projection heads on top of the text-encoded node and hyperedge representations. The projection heads map semantic embeddings into a contrastive space, preventing the contrastive loss from directly constraining the semantic representation space.
\begin{equation}
\begin{aligned}
     W^{(k)} = g_{\phi}(P^{(k)}), D^{(k)}=g_{\psi}(Q^{(k)}),
\end{aligned}
\end{equation}
both $g_{\phi}$ and $g_{\psi}$ are implemented as two-layer multilayer perceptrons (MLPs) with ELU activation to introduce nonlinearity.

We then adopt three contrastive learning objectives: node-level contrast, which aims to distinguish the representation of the same node across two augmented views from those of other nodes; group-level contrast, which differentiates the representations of the same hyperedge (area-level semantics) across two views from those of other hyperedges; and all-level contrast, which seeks to discriminate “real” node–hyperedge memberships from “fake” ones across different views.
In this work, we employ the InfoNCE loss~\cite{oord2018representation} as the contrastive learning objective.

For node-level contrast, given any node $w_{i}$, its representation in the first view, $w_{i}^{(1)}$, is treated as the anchor, while the corresponding representation in the second view, $w_{i}^{(2)}$, is regarded as the postive sample. The representations of all other nodes are considered negative samples. In this work, we adopt cosine similarity as the scoring function, defined as $s(u,v) = \frac{u^{T}v}{||u||||v||}$. Then the loss function for each positive node pair is defined as:

\begin{equation}
\small
\begin{aligned}
    \mathbf{l}_{n} (w_{i}^{(1)}, w_{i}^{(2)}) = -\log\frac{\text{exp}(s(w_{i}^{(1)}, w_{i}^{(2)})/\tau_{n})}{\sum_{k=1}^{|N|}\text{exp}(s(w_{i}^{(1)}, w_{k}^{(2)})/\tau_{n})},
\end{aligned}
\end{equation}
where $w_{i}^{(2)}$ denotes the representations of other nodes in the second view with $k \neq i$, and $\tau_{n}$ is a temperature parameter. We then symmetrize this loss function and compute the average over all positive sample pairs, as shown below:
\begin{equation}
\small
\begin{aligned}
    \mathcal{L}_{n} = \frac{1}{2|N|} \sum_{i=1}^{|N|}(\mathbf{l}_{n}(w_{i}^{(1)},w_{i}^{(2)})+\mathbf{l}_{n}(w_{i}^{(2)},w_{i}^{(1)})).
\end{aligned}
\end{equation}


For area-level contrast, given any hyperedge $z_{j}$, the representation $d_{j}^{(1)}$ from the first view is treated as the anchor, while the corresponding representation $d_{j}^{(2)}$ from the other view is regarded as the positive sample. The representations of all remaining hyperedges are considered negative samples. The loss for each positive hyperedge pair is defined as follows:
\begin{equation}
\small
\begin{aligned}
    \mathbf{l}_{g} (d_{j}^{(1)}, d_{j}^{(2)}) = -\log\frac{\text{exp}(s(d_{j}^{(1)}, d_{j}^{(2)})/\tau_{g})}{\sum_{k=1}^{|K|}\text{exp}(s(d_{j}^{(1)}, d_{j}^{(2)})/\tau_{g})},
\end{aligned}
\end{equation}
where $d_{j}^{(2)}$ denotes the representations of other hyperedges in the second view with $k \neq j$, and $\tau_{g}$ is a temperature parameter. We then symmetrize this loss function and compute the average over all positive sample pairs, as shown below:
\begin{equation}
\small
\begin{aligned}
    \mathcal{L}_{g} = \frac{1}{2|K|} \sum_{j=1}^{|K|}(\mathbf{l}_{g}(d_{j}^{(1)},d_{j}^{(2)})+\mathbf{l}_{g}(d_{j}^{(2)},d_{j}^{(1)})).
\end{aligned}
\end{equation}


For all-level contrast, the node representation $w_{i}^{(1)}$ from the first view is treated as the anchor, while the hyperedge representation $d_{j}^{(2)}$ from the other view that has a membership relation with the node is regarded as the positive sample. Negative samples are drawn from the representations of other hyperedges that are not associated with the node. To distinguish “real” node–hyperedge memberships from “fake” ones, we introduce a discriminator $\mathcal{D}: \mathbb{R}^{F^{'}} \times \mathbb{R}^{F^{''}} \rightarrow  \mathbb{R}$, which outputs a probability score for a given node–hyperedge representation pair. We optimize the following objective function:
\begin{equation}
\small
\begin{aligned}
    \mathbf{l}_{m} = -\log\frac{\text{exp}(D(w_{i}, d_{j}/\tau_{m})}{\text{exp}(D(w_{i}, d_{j}/\tau_{m})+\sum_{j^{'}\in\mathcal{N}_{i}^{-}}\text{exp}(D(w_{i}^{(1)}, d_{j^{'}})/\tau_{m})},
\end{aligned}
\end{equation}
where the negative sample set $ \mathcal{N}_{i}^{-}$ is defined as: $\mathcal{N}_{i}^{-} = \{j^{'}|h_{ij^{'}=0}\}$, where $\tau_{m}$ is a temperature parameter. For each node–hyperedge membership, two node–hyperedge representation pairs can be obtained from the two symmetric views. The final objective function for all-level contrast is defined as follows:
\begin{equation}
\small
\begin{aligned}
    \mathcal{L}_{m} = \frac{1}{2|K|} \sum_{i=1}^{|N|}\sum_{j=1}^{|K|}\mathbb{I}[h_{ij}=1](\mathbf{l}_{m}(w_{i}^{(1)},d_{j}^{(2)})+\mathbf{l}_{m}(w_{i}^{(2)},d_{j}^{(1)})).
\end{aligned}
\end{equation}

Finally, contrastive loss is formulated as:
\begin{equation}
\begin{aligned}
    \mathcal{L} = \mathcal{L}_{n} + \alpha_{g}\mathcal{L}_{g} + \alpha_{m}\mathcal{L}_{m},
\end{aligned}
\end{equation}
where $\alpha_{g}$ and $\alpha_{m}$ are the weights of $\mathcal{L}_{g}$ and $\mathcal{L}_{m}$, respectively. In all experiments, we simply set $\alpha_g = \alpha_m = 1$.

Based on this strategy, we further refine the hypergraph construction process, resulting in semantic hypergraph knowledge that serves as structured input for the VLM.

\begin{table*}[!t]
\centering
\renewcommand\arraystretch{1.15}
\setlength{\tabcolsep}{0.08cm}
\label{tab:metrics_comparison}
\scalebox{1}{
\begin{tabular}{lccccccccccc}
\specialrule{1.5pt}{0pt}{0pt}
Method & Type & Size & \multicolumn{3}{c}{ActPlan-1K(ctrf.)} & \multicolumn{3}{c}{ActPlan-1K(norm.)} & \multicolumn{3}{c}{ALFRED} \\
\cmidrule(lr){4-6} \cmidrule(lr){7-9} \cmidrule(lr){10-12}
 & & & Exec.$\uparrow$ & LCS$\uparrow$ & Corr.$\uparrow$ & Exec.$\uparrow$ & LCS$\uparrow$ & Corr.$\uparrow$ & Exec.$\uparrow$ & LCS$\uparrow$ & Corr.$\uparrow$ \\
\specialrule{0.5pt}{0pt}{0pt}

GPT-4o & Close-set & - & 49.2 & 0.48 & 21.4 & 56.5 & 0.59 & 39.8 & \textbf{90.4} & 0.60 & 47.4 \\
Gemini-Pro-1.5 & Close-set & - & 46.0 & 0.51 & 25.8 & 51.0 & 0.55 & 38.6 & 84.0 & 0.60 & 46.6 \\
\specialrule{0.5pt}{0pt}{0pt}

LLaVa-OV & Open-set & 7B & 37.3 & 0.49 & 24.6 & 50.6 & 0.55 & 35.5 & 78.2 & 0.42 & 29.8 \\
VideoLLaMA2 & Open-set & 7B & 30.2 & 0.43 & 20.3 & 40.6 & 0.48 & 31.9 & 73.1 & 0.57 & 43.9 \\
DeepSeek-VL2 & Open-set & 4.5B & 34.9 & 0.44 & 23.8 & 38.2 & 0.50 & 24.4 & 72.1 & 0.54 & 32.1 \\
Qwen2-VL & Open-set & 7B & 43.7 & 0.53 & 27.9 & 57.8 & 0.59 & 37.9 & 81.6 & 0.56 & 43.8 \\
InternVL2 & Open-set & 8B & 44.9 & 0.52 & 26.1 & 52.8 & 0.57 & 38.2 & 81.9 & 0.58 & 45.3 \\
InternVL2 (Pla) & Open-set & 8B & 48.2 & 0.53 & 30.2 & 53.9 & 0.58 & 39.0 & 81.8 & 0.58 & 45.7 \\
InternVL3 & Open-set & 8B & 45.2 & 0.53 & 28.2 & 52.9 & 0.57 & 38.4 & 81.9 & 0.58 & 45.5 \\
InternVL3.5 & Open-set & 8B & 45.1 & 0.52 & 27.8 & 53.1 & 0.58 & 38.5 & 82.0 & 0.59 & 45.6 \\
\specialrule{0.5pt}{0pt}{0pt}

Embodied-GPT & Specialized & 7B & 39.8 & 0.48 & 21.5 & 54.3 & 0.57 & 36.6 & 70.7 & 0.56 & 41.6 \\
LLaPa & Specialized & 8B & 53.2 & 0.57 & 36.1 & 62.9 & 0.62 & 45.2 & 85.3 & 0.62 & 48.6 \\
\rowcolor{gray!20} GaLa\textbf{(ours)} & Specialized & 8B & \textbf{55.1} & \textbf{0.59} & \textbf{37.2} & \textbf{65.8} & \textbf{0.66} & \textbf{47.5} & 89.9 & \textbf{0.67} & \textbf{50.1} \\

\specialrule{1.5pt}{0pt}{0pt}
\end{tabular}}
\caption{The table reports a quantitative comparison of different models on the ActPlan-1K dataset (including both counterfactual and normal activities) and the ALFRED dataset.}
\end{table*}

\section{Experiments}

\subsection{Experimental Setup}
\paragraph{Dataset.} To evaluate the model performance, we conducted experiments on two benchmarks: ActPlan-1K~\cite{su2024actplan} and ALFRED~\cite{shridhar2020alfred}. ActPlan-1K is currently the only multimodal planning dataset that supports counterfactual reasoning, used to examine the model’s adaptive planning capability under unconventional conditions. ALFRED is employed to test the model’s zero-shot generalization performance in embodied tasks. More detailed information about the dataset can be found in the Appendix ~\ref{dataset}.

\begin{figure}[t]
  \includegraphics[width=\linewidth]{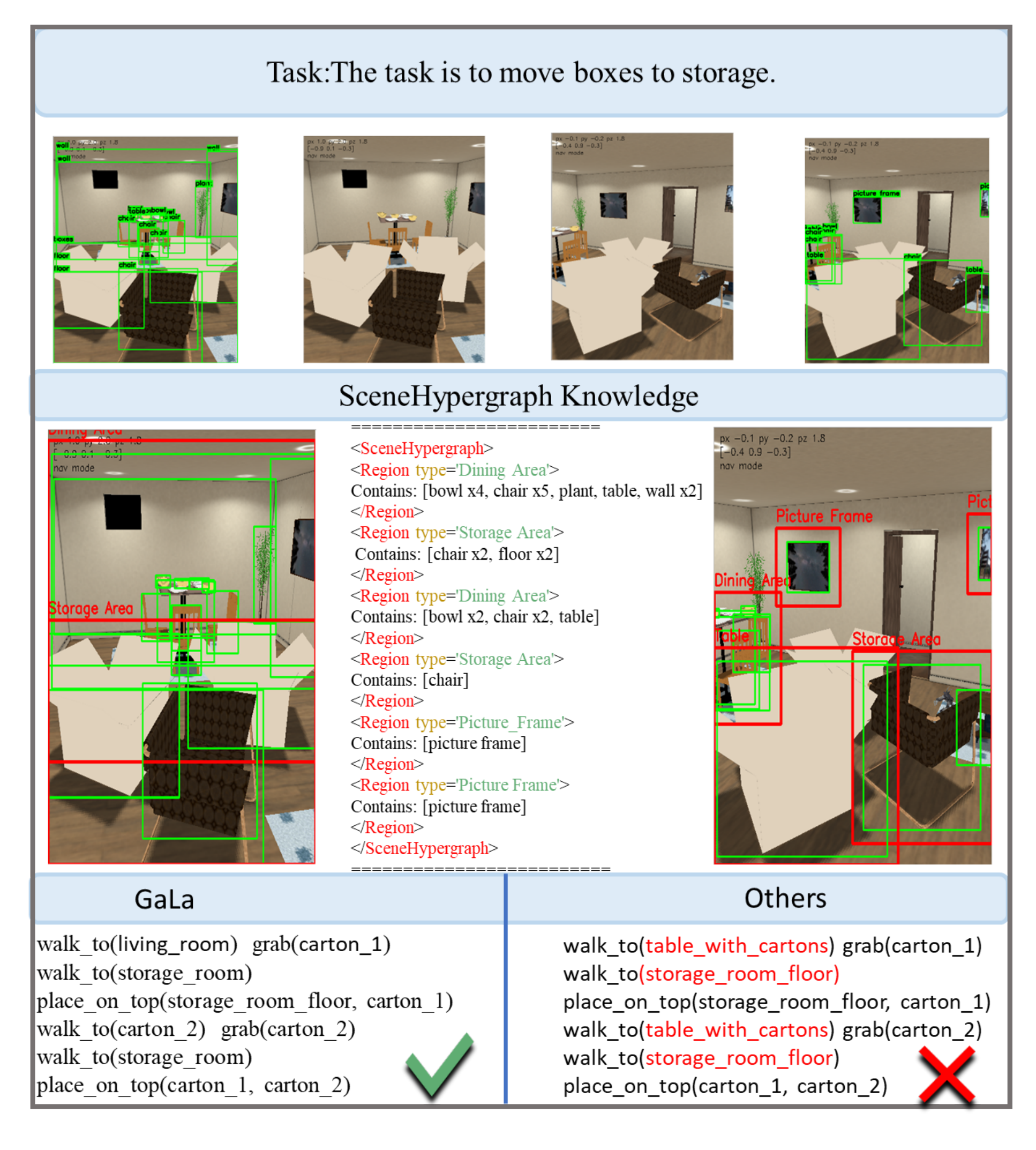}
  \caption{This figure illustrates the instruction decomposition process of GaLa.}
  \label{fig:qua}
\end{figure}

\paragraph{Baselines.} We compare the GaLa model with a wide range of state-of-the-art baseline methods, which are divided into three categories. The first category consists of close-set VLM-based model methods.: GPT-4o~\cite{hurst2024gpt}; Gemini-Pro-1.5~\cite{team2023gemini}. The second category consists of open-set VLM-based model methods: LLaVa-OV~\cite{li2024llava}; VideoLLaMA2~\cite{cheng2024videollama}; DeepSeek-VL2~\cite{wu2024deepseek}; Qwen2-VL~\cite{wang2024qwen2}; InternVL2~\cite{chen2024far}; InternVLwithPlasma’s~\cite{brahman2024plasma}; InternVL3~\cite{zhu2025internvl3}; InternVL3.5~\cite{wang2025internvl3}. The third category consists of specialized procedural planning model methods: Embodied-GPT~\cite{mu2023embodiedgpt}; LLaPa~\cite{sun2025llapa}. We uniformly set the maximum generation length to 512 tokens.

\paragraph{Implementation Details.} Our model GaLa uses InternVL3-8B as its backbone, where the visual encoder adopted the InternViT architecture, and the language model was based on InternVL3. Our model is trained for 30 epochs on benchmark hardware (Intel Xeon 8558@4.00GHz, NVIDIA HGX H200 141G, Ubuntu22.04). Training is conducted using the Adam optimizer with an initial learning rate of 2e-5, no weight decay, and a batch size of 128. More detailed information about the implementation details can be found in Appendix~\ref{details}.

\paragraph{Evaluation Metrics.} 
To comprehensively evaluate the quality of the generated action sequences, this study adopts a multidimensional evaluation framework that integrates three key aspects: the executability of the plan, its sequential similarity to reference plans, and the task correctness. This aligns with common practices in the field~\cite{ puig2018virtualhome, song2023llm, brahman2024plasma}. Specifically: (i) Executability quantifies the proportion of generated actions that can be successfully executed in the simulated environment; (ii) Sequential similarity is measured by calculating the Longest Common Subsequence (LCS) between the generated plan and human-annotated references; (iii) Task correctness assesses whether the action sequence ultimately accomplishes the predefined task objective.

\subsection{Overall Performance}



\paragraph{Quantitative Results.} As shown in Table~\ref{tab:metrics_comparison}, GaLa is trained on the ActPlan-1K and ALFRED datasets and evaluated using three metrics. On the ActPlan-1K dataset, for counterfactual activities, GaLa achieves 55.1\% Executability, 0.59 LCS, and 37.2\% Correctness. For normal activities, it attains 65.8\% Executability, 0.66 LCS, and 47.5\% Correctness. We observe that the performance gains on normal activities are larger than those on counterfactual activities, indicating that counterfactual tasks are inherently more challenging. Compared with its baselines, GaLa achieves substantial improvements across all evaluation metrics. On the ALFRED dataset, GaLa reaches 89.9\% Executability, 0.67 LCS, and 50.1\% Correctness. 



\begin{table}[!t]
\centering
\renewcommand\arraystretch{1}
\setlength{\tabcolsep}{0.02cm}

\scalebox{1}{
\begin{tabular}{lcccccc}
\specialrule{1.5pt}{0pt}{0pt}
Method & \multicolumn{3}{c}{ActPlan-1K(ctrf.)} & \multicolumn{3}{c}{ActPlan-1K(norm.)} \\ 
\cmidrule(lr){2-4} \cmidrule(lr){5-7}
 & Exec.$\uparrow$ & LCS$\uparrow$ & Corr.$\uparrow$ & Exec.$\uparrow$ & LCS$\uparrow$ & Corr.$\uparrow$ \\
\specialrule{0.5pt}{0pt}{0pt}

GaLa & \textbf{55.1} & \textbf{0.59} & \textbf{37.2} & \textbf{65.8} & \textbf{0.66} & \textbf{47.5} \\
w/o Hyper & 51.7 & 0.56 & 32.2 & 61.5 & 0.62 & 44.3 \\
w/o $H_{\mathcal{L}_{n}}$ & 52.1 & 0.56 & 33.6 & 62.3 & 0.63 & 45.1 \\
w/o $H_{\mathcal{L}_{g}}$ & 52.7 & 0.57 & 34.0 & 63.2 & 0.64 & 45.6 \\
w/o $H_{\mathcal{L}_{m}}$ & 52.4 & 0.57 & 33.9 & 63.5 & 0.64 & 46.1 \\

\specialrule{1.5pt}{0pt}{0pt}
\end{tabular}
}
\caption{Ablation Results of GaLa on ActPlan-1K.}
\label{tab:ab1}
\end{table}

These results validate that explicitly modeling implicit spatial relations and object-cluster semantics through hypergraphs can effectively guide VLMs to generate more spatially aware and accurate action sequence

\paragraph{Qualitative Results.} Figure~\ref{fig:qua} and~\ref{fig:apex1}  present a case study of the GaLa model, illustrating how it constructs hypergraph-based knowledge from images and ultimately generates a correct action sequence. The figures also compares GaLa with other models that do not incorporate hypergraph construction, highlighting the differences in the generated action sequences.

\subsection{Ablation Study.} To systematically validate the design choices of GaLa, we conduct a series of ablation studies to evaluate the effectiveness of the hypergraph information and the Tri-View Hypergraph mechanism.

As shown in Table~\ref{tab:ab1}, we analyze the impact of incorporating hypergraph construction information (w/o Hyper) into this domain on procedural planning performance. We also conduct ablation studies to examine the effects of introducing hypergraph-based contrastive learning on nodes ($H_{\mathcal{L}_{n}}$), edges ($H_{\mathcal{L}_{g}}$), and both nodes and edges ($H_{\mathcal{L}_{m}}$), respectively.



\begin{figure}[t]
  \includegraphics[width=1\linewidth]{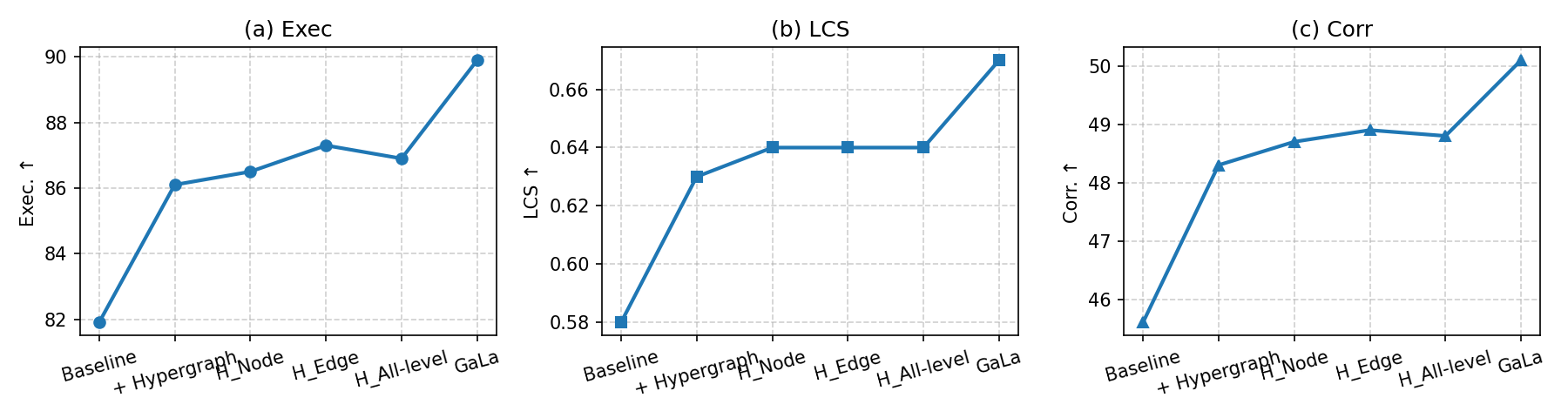}
  \caption{Ablation studies of GaLa on the ALFRED dataset in terms of Exec., LCS, and Corr. metrics.}
  \label{fig:abl2}
\end{figure}

Through ablation studies on the ActPlan-1K dataset, we find that introducing the hyperedge-based graph construction strategy (w/o Hyper) into the procedural planning framework effectively models implicit relations among objects and enables the extraction of object-cluster attributes, thereby substantially improving the performance of VLM-based models. Under counterfactual activity settings, the Exec. score increases from 45.2\% to 51.7\%, the LCS score improves from 0.53 to 0.56, and the Corr. score rises from 28.2\% to 34.2\%. For normal activities, similar gains are observed, with the Exec. score improving from 52.9\% to 61.5\%, the LCS score increasing from 0.57 to 0.62, and the Corr. score rising from 38.4\% to 44.3\%.

Moreover, building upon the constructed hypergraph, the incorporation of the Tri-view strategy consistently yields further improvements across all three evaluation metrics, demonstrating its effectiveness in enhancing structured semantic representations.

As shown in Figure~\ref{fig:abl2}, we conduct corresponding ablation experiments on the ALFRED dataset. The results demonstrate that incorporating the hypergraph-based strategy effectively improves the model’s performance across all evaluation metrics. Furthermore, introducing the three contrastive learning objectives leads to additional and consistent performance gains.
\section{Conclusion}
We presented GaLa, a Graph-augmented vision–language framework for multimodal procedural planning in embodied AI. By introducing a hypergraph-based intermediate representation, GaLa explicitly models implicit spatial relations and region-level semantic structures that are often overlooked by existing VLM-based approaches. Furthermore, the proposed Tri-View HyperGraph Encoder enforces semantic consistency across node-level, area-level, and all-level views via contrastive learning, enabling effective injection of structured scene knowledge into downstream reasoning. Experimental results on ActPlan-1K and ALFRED demonstrate that GaLa significantly improves execution success, plan consistency, and correctness, validating the effectiveness of explicitly modeling implicit semantic structures for robust procedural planning.
\section{Acknowledgments}
This work was supported by the National Natural Science Foundation of China (Grant No. 62277011), Project of Chongqing MEITC(Grant No. YJX-2025001001009) , Open Research Fund from Guangdong Laboratory of Artificial Intelligence and Digital Economy (SZ) (Grant No.GML-KF-24-18) and CAAI-CANN Open Fund, developed on OpenI Community.

\section{Limitations}

Although GaLa achieves substantial performance improvements over InternVL3 on both the ActPlan-1K and ALFRED benchmarks, it still has several limitations. First, GaLa improves model performance by constructing and refining hypergraphs from visual inputs; however, since the ActPlan-1K dataset provides only a limited set of images associated with each action, the constructed hypergraphs may be incomplete, potentially leading to the loss of important semantic information. Second, while GaLa represents an initial attempt to introduce hypergraph-based modeling into the procedural planning domain and demonstrates consistent gains over baseline methods, it does not fully explore the potential of more advanced graph-theoretic techniques that could further enhance model performance.



\bibliography{custom}
\appendix

\newpage
\section{Appendix: Method}
\label{sec:appendix:method}

\subsection{Prompt}
\begin{figure}
  \includegraphics[width=\linewidth]{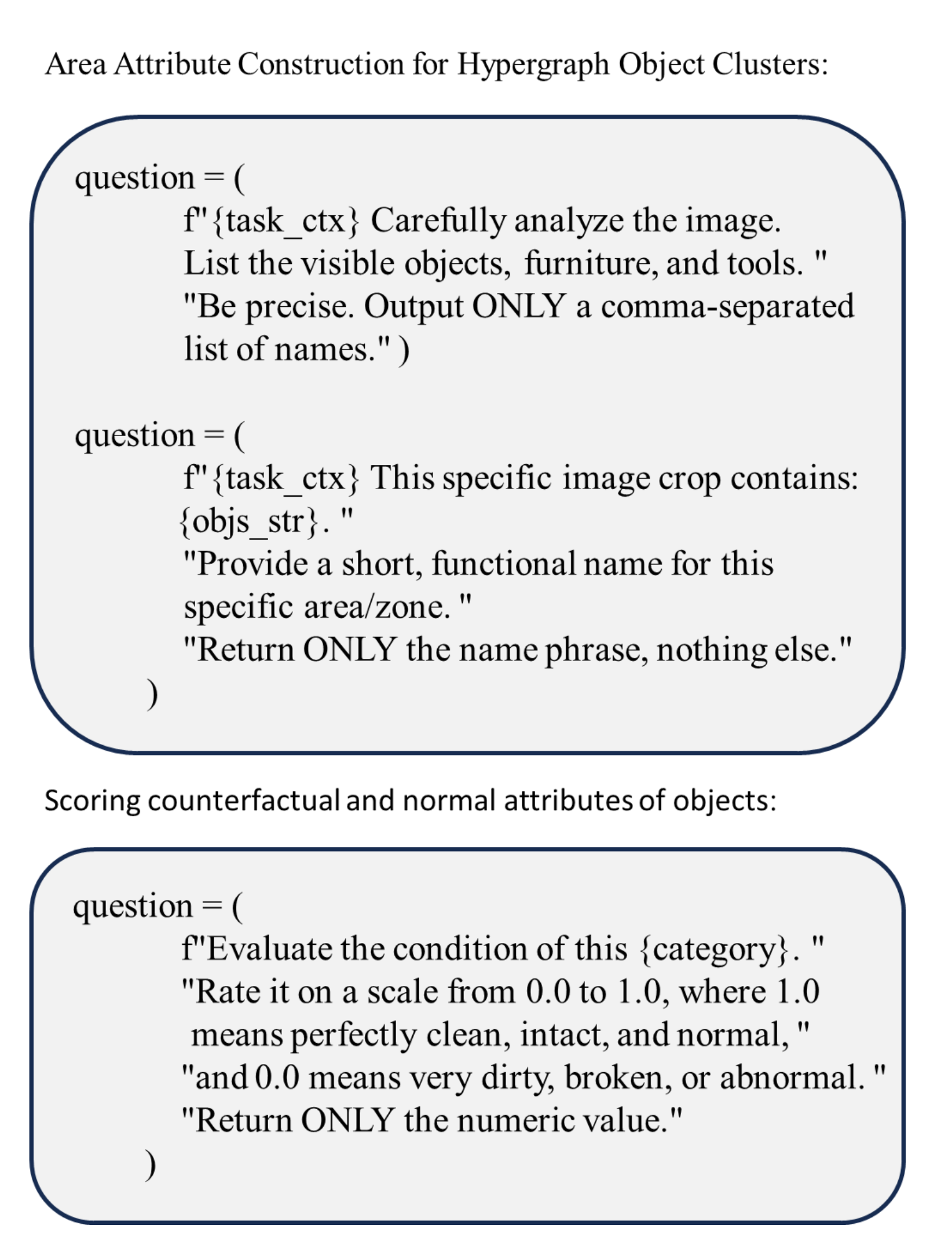}
  \caption{This figure illustrates the instruction decomposition process of GaLa.}
  \label{fig:pro}
\end{figure}

As shown in Figure~\ref{fig:pro}, we first apply YOLO-World~\cite{cheng2024yolo} to detect objects in the image and extract their attributes, based on which a hypergraph is constructed. We then feed each object cluster together with a predefined prompt into a large language model to generate semantically meaningful area-level attributes, enabling the model to further uncover hierarchical region-level relationships within the image.

For the ActPlan-1K dataset, we additionally design a specific prompt to send individual object attributes to the language model, which outputs a scalar value in the range [0,1] to indicate the degree of normality of each attribute.


\section{Appendix: Experiments}
\label{sec:appendix:exp}

\subsection{Dataset}
\label{dataset}
The main components of the ActPlan-1K dataset are located in the annotation directory. This directory contains multiple scenes, and each scene includes several tasks. For each task, there are both counterfactual activities and normal activities. Each activity consists of multiple samples that share the same task description. Each sample contains several scene-frame images and two gold files with different styles. The ActPlan-1K dataset includes 703 training samples, 243 test samples, and 243 validation samples.

In our experiments, we use the Lite version of the ALFRED dataset, which includes task descriptions and scene information. The dataset also provides high-level actions in PDDL format. By running the provided scripts, image frames corresponding to the task execution process can be generated. To align with our model, we manually select a small number of key scene images based on the actions in the dataset, allowing most viewpoints of the scene to be covered with fewer images. The ALFRED dataset contains 2,435 training samples, 483 test samples, and 242 validation samples.
\subsection{Implementation Details.}

\label{details}
The GaLa model is trained for 30 epochs on 8 NVIDIA HGX H200 (141 GB) GPUs. Training is conducted using the Adam optimizer with an initial learning rate of 2e-5, no weight decay, and a batch size of 128. The original input images have a resolution of 600 × 600, which is forcibly resized to 448 × 448 before being fed into the model.

The TriCL module consists of 2 layers with a projection dimension of 512. The InfoNCE temperature is set to 0.07, and the TriCL loss weight is 0.5. The visual encoder follows the InternViT architecture, with a ViT patch size of 14, while the text encoder is Qwen3-8B.
\subsection{Case Study}

\begin{figure*}
  \includegraphics[width=\linewidth]{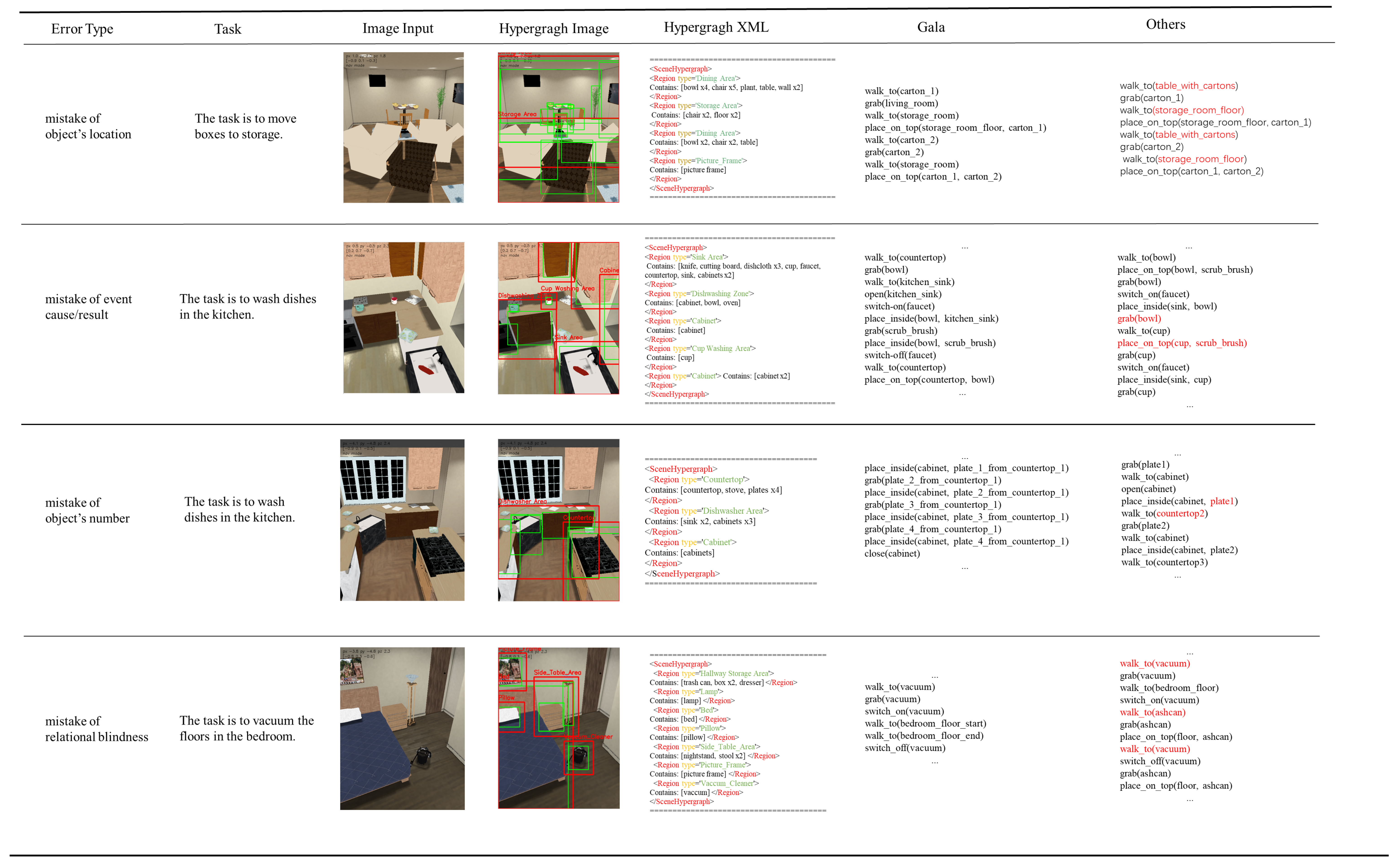}
  \caption{A comparison of action instruction predictions between the GaLa model and other models.}
  \label{fig:apex1}
\end{figure*}

As shown in Figure~\ref{fig:apex1}, this figure illustrates the overall hypergraph construction process of the GaLa model as well as typical errors that may occur in other models. The first column presents different types of errors commonly made by existing approaches, including mistakes of object location, mistakes of event cause, mistakes of object number, and mistakes of relational blindness. The second column contains the corresponding instructional task descriptions. The third column shows the input images, while the fourth column visualizes the constructed hypergraphs. The fifth column provides the XML-formatted representation of the constructed hypergraph. The last two columns compare the action predictions generated by GaLa and by other baseline models.

From these examples, we observe that constructing a hypergraph not only improves the overall performance of the model, but also enables more effective mining of implicit spatial relations and deep semantic information from multimodal inputs. As a result, GaLa can alleviate logical deadlocks in action sequence prediction—particularly those caused by relational blindness—to a considerable extent.

\label{Example}

\end{document}